\documentclass[letterpaper]{article} 
\usepackage{aaai23}  
\usepackage{times}  
\usepackage{helvet}  
\usepackage{courier}  
\usepackage[hyphens]{url}  
\usepackage{graphicx} 
\usepackage{natbib}  
\usepackage{caption} 
\usepackage{subcaption}
\usepackage{algorithm}
\usepackage{algorithmic}

\usepackage{bbold}
\usepackage[normalem]{ulem}
\useunder{\uline}{\ul}{}
\usepackage{color}

\usepackage{amsfonts,amssymb,amsmath}
\usepackage{cleveref}  
\usepackage{multirow}
\usepackage{booktabs}
\usepackage{makecell}
\usepackage{balance}

\newtheorem{theorem}{Theorem}
\newtheorem{definition}{Definition}
\newtheorem{lemma}{Lemma}
\newtheorem{example}{Example}

\newcommand{\hgnn}{{M_G}}
\newcommand{\explain}[1]{X_{#1}}
\newcommand{\infl}{\textsf{IF}}
\newcommand{\inedge}{\textsf{InE}}
\newcommand{\outedge}{\textsf{OutE}}
\newcommand{\proxy}{{\rm proxy}}
\newcommand{\modelname}{xPath~} 
\newcommand{\modelns}{xPath} 
\newcommand{\eat}[1]{}
\newcommand{\std}[1]{{$\pm$#1}}

\usepackage{newfloat}
\usepackage{listings}
\DeclareCaptionStyle{ruled}{labelfont=normalfont,labelsep=colon,strut=off} 
\floatstyle{ruled}
\newfloat{listing}{tb}{lst}{}
\floatname{listing}{Listing}
\title{Towards Fine-Grained Explainability for Heterogeneous Graph Neural Network}
\author{
    Tong Li\textsuperscript{\rm 1},
    Jiale Deng\textsuperscript{\rm 1},
    Yanyan Shen\textsuperscript{\rm 1}\thanks{Corresponding author.},
    Luyu Qiu\textsuperscript{\rm 2},
    Yongxiang Huang\textsuperscript{\rm 2},
    Caleb Chen Cao\textsuperscript{\rm 2}
}
\affiliations{

    \textsuperscript{\rm 1} Shanghai Jiao Tong University\\
    \textsuperscript{\rm 2} Huawei Research Hong Kong\\
    \{2017lt, jialedeng, shenyy\}@sjtu.edu.cn,
    \{qiuluyu, huang.yongxiang2, caleb.cao\}@huawei.com 
}

\usepackage{bibentry}

\begin{document}

\maketitle
\begin{abstract}

	Heterogeneous graph neural networks (HGNs) are prominent approaches to node classification tasks on heterogeneous graphs. Despite the superior performance, insights about the predictions made from HGNs are obscure to humans. Existing explainability techniques are mainly proposed for GNNs on homogeneous graphs. They focus on highlighting salient graph objects to the predictions whereas the problem of how these objects affect the predictions remains unsolved. Given heterogeneous graphs with complex structures and rich semantics, it is imperative that salient objects can be accompanied with their influence paths to the predictions, unveiling the reasoning process of HGNs. In this paper, we develop xPath, a new framework that provides fine-grained explanations for black-box HGNs specifying a cause node with its influence path to the target node. In xPath, we differentiate the influence of a node on the prediction w.r.t. every individual influence path, and measure the influence by perturbing graph structure via a novel graph rewiring algorithm. Furthermore, we introduce a greedy search algorithm to find the most influential fine-grained explanations efficiently. Empirical results on various HGNs and heterogeneous graphs show that xPath yields faithful explanations efficiently, outperforming the adaptations of advanced GNN explanation approaches.  

\eat{
In parallel with the proliferation of heterogeneous graph neural networks(HGNs), understanding the reasons behind the predictions is urgently demanded in order to build trust and confidence in the models. Some researches develop interpretable HGNs and provide model-specific explanations, which fall short when access to the architecture or parameters of an HGN is unauthorized. 
Alternatively, we can adapt advanced model-agnostic explanation methods for GNN, which attribute the prediction to salient graph objects. However, they fail to unveil the reasoning logic of the model due to the complex semantics of heterogeneous graphs. 
In this paper, we propose to explain HGNs by fine-grained explanations, each of which consists of an influential node and a path that specifies how the node affects the prediction.  
We develop a novel graph rewiring algorithm to perturb the walks associated with the path and quantify the influence of a fine-grained explanation. We also design a greedy search algorithm to find the top-$K$ most influential fine-grained explanations efficiently. 
Extensive experiments on various HGNs and real-world heterogeneous graphs demonstrate \modelname outperforms the adaptations of the state-of-the-art GNN explainers. To the best of our knowledge, this work is the first explainability approach for black-box HGNs.
}
\end{abstract}

\section{Introduction}


Real-world graphs often come with nodes and edges in multiple types, which are known as heterogeneous graphs. Recently, heterogeneous graph neural networks (HGNs) have become one of the standard paradigms for modeling rich semantics of heterogeneous graphs in various 
application domains such as e-commerce, finance, and healthcare~\cite{lv2021we,wang2022survey}.
%
In parallel with the proliferation of HGNs, understanding the reasons behind the predictions from HGNs is urgently demanded in order to build trust and confidence in the models for both users and stakeholders. For example, a customer would be satisfied if an HGN-based recommender system accompanies recommended items with explanations; a bank manager may want to know why an HGN flagged an account as fraudulent.

To equip HGNs with the capability of providing explanations, some researches focus on developing interpretable models that use model parameters~\cite{li2021higher}, gradients~\cite{pope2019explainability, baldassarre2019explainability, schnake2021higher} or attention scores~\cite{hu2020heterogeneous,yang2021interpretable} to find salient information for the predictions. Unfortunately, they are model-specific and fall short when accesses to the architecture or parameters of an HGN are unauthorized, especially under confidentiality and security concerns~\cite{dou2020enhancing, liu2021pick}. 

Alternatively, we can adapt advanced model-agnostic explanation methods~\cite{yuan2020explainability} proposed for GNNs (on homogeneous graphs) to HGNs. These methods attribute model predictions to graph objects, such as nodes~\cite{vu2020pgm}, edges~\cite{ying2019gnnexplainer, luo2020parameterized, schlichtkrull2020interpreting, wang2021towards, lin2021generative} and subgraphs~\cite{yuan2021explainability}. Their goal is to learn or search for optimal graph objects that maximize mutual information with the predictions. While such explanations answer the question ``\emph{what} is salient to the prediction'', they fail to unveil ``\emph{how} the salient objects affect the prediction''. In particular, 
there may exist multiple paths in the graph to propagate the information of the salient objects to the target object and affect its prediction. Without distinguishing these different influential paths, the answer to the ``how'' question remains unclear, 
which could compromise the utility of the explanation. This issue becomes more prominent when it comes to explaining HGNs due to the complex semantics of heterogeneous graphs.  

\begin{figure}[t]
		\centering
		\includegraphics[width=1\columnwidth]{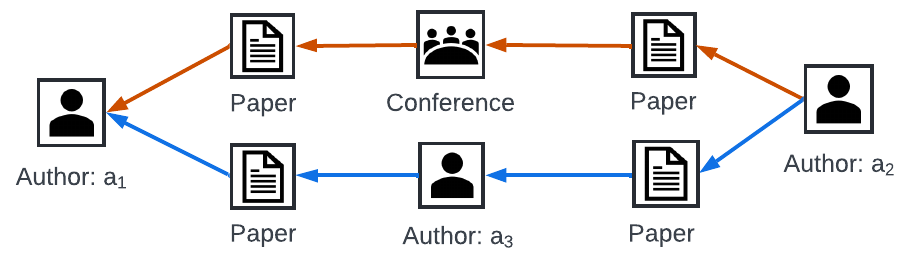}
		\caption{An explainer identifies author $a_2$ as the explanation for the prediction of author $a_1$. There are two semantically different paths showing how $a_2$ affects the prediction of $a_1$. }
		\label{fig:intro}
	\end{figure}

\begin{example}\label{ex:ex1}
    Consider an HGN that performs node classification on a heterogeneous academic graph with three types of nodes, \textsf{Author(A), Paper(P), Conference(C)}, as shown in Figure \ref{fig:intro}. The model classifies the research area of author $a_1$ as ``AI''. An explainer identified author $a_2$ as the explanation for the prediction of $a_1$. 
	To explore how $a_2$ affects the prediction, we notice there are two simple paths $P_1,P_2$ in the graph that relate $a_2$ to $a_1$. $P_1$ indicates $a_1,a_2$ have co-authored with $a_3$ individually
	and $P_2$ shows $a_1,a_2$ have published papers in the same conference $c$. 
	Assume the HGN can propagate $a_2$'s information to $a_1$ along both paths to affect the prediction.
	%
	The two paths carry different semantic meanings following \textsf{APAPA} and \textsf{APCPA} at the meta level respectively. They thus provide semantically different answers to the question ``how $a_2$ affects the prediction of $a_1$ by the HGN''.
	Without distinguishing the influence paths from $a_2$ (cause) to $a_1$ (effect), it is difficult to understand the reasoning process of the model (the ``how'' question).
\end{example}

The goal of this paper is to provide explanations for black-box HGNs that can answer ``what'' and ``how'' questions on node classification tasks.
Consider a heterogeneous graph $G$ with node set $V$ and edge set $E$ and an HGN $\hgnn$ that predicts the label of a target node $v_t\in V$ to be $\hgnn(v_t)$.
To achieve the goal, we propose to explain $\hgnn(v_t)$ with \emph{fine-grained explanations} in the form of $(v, P)$, where $v\in V$ is the cause of the prediction and $P$ is a simple path in the graph connecting $v$ to $v_t$ that unveils how $v$ affects the prediction. 
In Example~\ref{ex:ex1}, both $(a_2, P_1)$ and $(a_2, P_2)$ are fine-grained explanations for $\hgnn(a_1)$, corresponding to two different ways that $a_2$ affects the prediction of $a_1$.
Particularly, fine-grained explanations allow us to drill down the influence of a node on the prediction with respect to each individual influence path, and the node and path in such an explanation answer ``what'' and ``how'' questions respectively. 


To find fine-grained explanations that are faithful to $\hgnn(v_t)$, there are two challenging issues to be tackled. 
First, we need to measure the {influence} of any node $v$ on the prediction $\hgnn(v_t)$ with respect to a simple path $P$ from $v$ to $v_t$. 
A typical way is to perturb ($v$, $P$) and measure the change in the model prediction. However, traditional perturbations on graph data~\cite{ying2019gnnexplainer, vu2020pgm} like masking nodes or edges are inapplicable to our setting because multiple fine-grained explanations for $\hgnn(v_t)$ may share the same perturbed graph object and the prediction change caused by the perturbation is not specific to any of the explanations.
In Example~\ref{ex:ex1} , masking $a_2$ perturbs the influences of ($a_2$, $P_1$) and ($a_2$, $P_2$) on $\hgnn(a_1)$ simultaneously.
To address the issue, our idea is to perturb all the \emph{walks} that participate in propagating the information of $v$ to $v_t$ with respect to $P$, where a walk corresponds to the trajectory of a possible information flow in the HGN. 
We develop a novel graph rewiring algorithm such that all the walks facilitating the information of $v$ to flow to $v_t$ along $P$ are blocked and the other walks are not affected. 

Second, we are interested in highly influential fine-grained explanations on $\hgnn(v_t)$, i.e., the salient part for the prediction. However, a full enumeration of possible fine-grained explanations for $\hgnn(v_t)$ can be prohibitively expensive or unaffordable as the number of simple paths is exponential to the number of edges. 
We design an efficient search algorithm to explore the space of fine-grained explanations in a greedy manner. It generally follows the breadth-first search process but expands a small number of explanations with highest influence scores in each layer. The experiments show the search is able to find faithful fine-grained explanations with high efficiency. 

To summarize, this paper introduces an efficient fine-grained explanation framework named~\modelname for black-box HGNs on node classification tasks. 
The major contributions are as follows.
{\bf (1)} We propose a new fine-grained explanation scheme to explain black-box HGNs on node classification tasks. Each explanation involves a cause node to the prediction and a path specifies how the node affects the prediction. 
{\bf (2)} We develop a novel graph rewiring algorithm to perturb the walks associated with the path in a fine-grained explanation and quantify the influence of a node on the prediction w.r.t. each individual path to the target node. 
{\bf (3)} We design a greedy search algorithm to find top-$K$ most influential fine-grained explanations efficiently. 
{\bf (4)} Extensive experiments on various HGNs and real-world heterogeneous graphs demonstrate \modelname can find explanations that are faithful to HGNs and outperform the adaptations of the state-of-the-art GNN explanation methods.\footnote{Source code at https://github.com/LITONG99/xPath.}  
To the best of our knowledge, this work is the first explanation approach for black-box HGNs. The proposed fine-grained explanations distinguish influence paths with different semantics, allowing practitioners to understand the reasoning process of HGNs on complex heterogeneous graphs.

\section{Preliminaries}
\begin{definition}[Heterogeneous Graph $G$] A heterogeneous graph $G=(V, E, \phi, \psi)$ consists of a node set $V$ and a directed edge set $E$, where $\phi: V \rightarrow T $ is the node type mapping function and $\psi: E \rightarrow S $ is the edge type mapping function. $T$ and $S$ denote the respective sets of predefined node types and edge types, where $|T| + |S|>2$.
\end{definition}

\begin{definition}[Heterogeneous GNN (HGN) $\hgnn$] 
 Let $\hgnn: V\rightarrow C$ be a trained heterogeneous graph neural network model that performs node classification on the heterogeneous graph $G$, where $C$ is the label set. For any node $v_t\in V$, $\hgnn(v_t)$ is the model prediction. 
\end{definition}

In this paper, we assume $\hgnn$ performs message passing along the directed edges in $G$, which holds for most existing HGNs~\cite{schlichtkrull2018modeling, fu2020magnn, hu2020heterogeneous, lv2021we}. 
Apart from this assumption, we treat HGN as a black-box model without requiring the detailed model architecture and parameters. 

\begin{definition}[Fine-Grained Explanation $\explain{G}$]
	To explain a model prediction $\hgnn(v_t)$, we propose a fine-grained explanation $\explain{G}(v_t)$ in the form of $(v,P)$ where $v\in V\backslash\{v_t\}$ is regarded as the {cause} for the prediction and $P=\langle v, \cdots, v_t\rangle$ is a directed simple (acyclic) path connecting $v$ with $v_t$ (according to the edge set $E$) indicating the way in which $v$ affects $\hgnn(v_t)$. 
\end{definition}
Let $\mathcal{P}_G(v_t)$ denote all the simple paths in the graph ended with $v_t$. We can obtain the set of all the possible fine-grained explanations for $\hgnn(v_t)$ as: $\mathcal{X}_G(v_t)=\{\explain{G}(v_t)=(v,P) \mid P=\langle v,\cdots, v_t\rangle\in \mathcal{P}_G(v_t)\}$.
Among all the fine-grained explanations in $\mathcal{X}_G(v_t)$, we are interested in the most influential ones.
We denote by $\infl(\explain{G}(v_t))$ the \emph{influence} of node $v$ on the prediction $\hgnn(v_t)$ w.r.t. path $P$, which is also called the \emph{influence score} of $\explain{G}(v_t)$ for simplicity. 
\subsubsection{Problem Formulation.} For each prediction made by a black-box HGN, the goal of this paper is to find $K$ fine-grained explanations with the highest influence scores. 
\eat{
The first problem is to reasonably and efficiently compute $\infl(\explain{G}(v_t))$.
Then, for a specific prediction $\hgnn(v_t)$, the second problem is to efficiently generate a list of fine-grained explanations with top-$K$ influence score  $\mathcal{X}^K_G(v_t)\subset \mathcal{X}_G(v_t)$.
}

\section{Methodology}

In this section, we first introduce the notion of walk perturbation which is crucial for computing the $\infl(\explain{G}(v_t))$. We then present a novel graph rewiring algorithm for walk perturbation and discuss its properties. Finally, we describe our explanation framework \modelname that can find top-$K$ fine-grained explanations for each prediction efficiently. 

\subsection{Walk Perturbation}

We now consider a fine-grained explanation $\explain{G}(v_t)$ 
and want to measure $\infl(\explain{G}(v_t))$.
As described before, traditional perturbation techniques such as node masking and edge dropping fail to untangle the influences of multiple fine-grained explanations that have overlapping nodes.

\begin{figure}[t]
        \centering
        \begin{subfigure}[t]{0.48\columnwidth} \includegraphics[height=2.8cm]{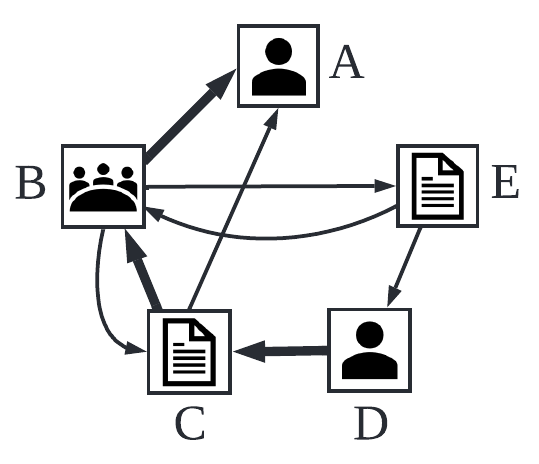}
            \caption{}
            \label{fig:flowset-1}
        \end{subfigure}
        \begin{subfigure}[t]{0.48\columnwidth}
            \includegraphics[height=2.8cm]{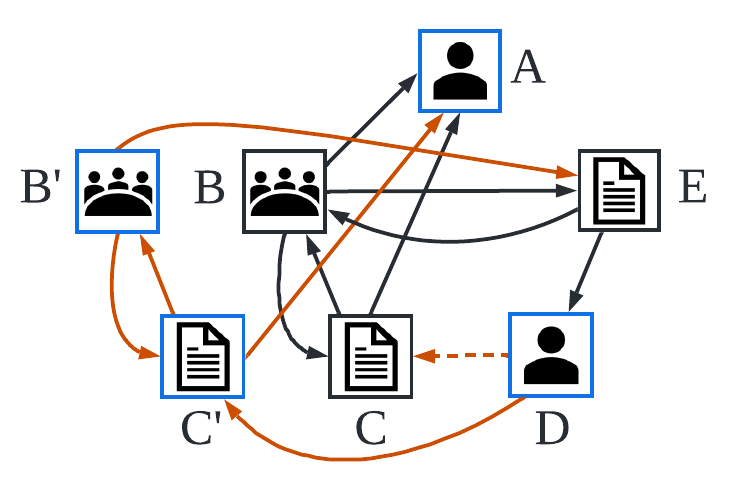}
            \caption{}
            \label{fig:flowset-2}
        \end{subfigure}
        \caption{(a) Illustration of the fine-grained explanation ($v=D$, $P=\langle D,C,B,A\rangle$)  for the prediction of $A$ in graph $G$. (b) The rewired graph $G_R^P$ with $B', C'$ as proxies.}
        \label{fig:flowset}
    \end{figure}
Our key insight is that the HGN model performs actual information propagation from $v$ to $v_t$ w.r.t. $P$ via \emph{walks} on the graph. 
Each walk is an ordered node sequence connected by edges in the graph showing the trajectory of a specific information flow. In Figure~\ref{fig:flowset-1}, two walks $\langle D,C,B,A \rangle$ and $\langle D,C,B,C,B,A \rangle$ indeed participates in propagating the information of $D$ to $A$ w.r.t. the path $\langle D,C,B,A \rangle$. 
On the contrary, the walk $\langle D,C,A \rangle$ does not follow the path to propagate information of $D$ to $A$. 
Intuitively, if we can identify all the walks associated with $P$ and block them to prevent the information of $v$ from flowing to $v_t$ w.r.t. $P$, the observed prediction change will imply the $\infl(\explain{G}(v_t))$.

We next formally define walks and specify the walks that are associated with a simple path $P$.

\begin{definition}[Walk $W_G$]
Given a heterogeneous graph $G$, a walk $W_G$ is an ordered node sequence $\langle v_1, \cdots, v_L \rangle$ where $\langle v_i,v_{i+1}\rangle\in E$ for any $i\in[1,L-1]$ and $L\geq 2$. Note that different from simple paths, a walk can involve (self-)loops.
\end{definition}

\begin{definition}[Walk $\mathcal{W}_G^P$ Associated With $P$]\label{def:associate}
	Given a fine-grained explanation $\explain{G}(v_t)=(v,P)$ for the prediction $\hgnn(v_t)$, a walk $W_G=\langle v_1, \cdots, v_L \rangle$ is associated with $P$ iff it includes a suffix $\langle v_l, \cdots, v_L\rangle$ satisfying the following three conditions:\\
	(i) $v_l=v\wedge v_L=v_t\wedge \forall l\leq i<L, v_i\neq v_t $; \\
	(ii) for all $i\in [l, L-1]$, we have $v_i=v_{i+1}$ or exactly one of the two edges $\langle v_i, v_{i+1} \rangle$,  $\langle v_{i+1}, v_{i} \rangle$ exists in the path $P$; \\
	(iii) after erasing all the loops in the suffix, we obtain exactly the same path as $P$. 
\end{definition}

We denote by $\mathcal{W}_G^P$ the set of all the walks associated with $P$ and have the following theorem.

\begin{theorem}\label{the:neq}
	Given two different fine-grained explanations $\explain{G,1}(v_t)=(v_1,P_1)$ and $\explain{G,2}(v_t)=(v_2,P_2)$ where $P_1 \neq P_2$ for the prediction $\hgnn(v_t)$, we have $\mathcal{W}_G^{P_1}\neq  \mathcal{W}_G^{P_2}$.  
\end{theorem}

Based on Theorem~\ref{the:neq}, we can distinguish the influence of $v$ w.r.t. different simple paths connecting $v$ to $v_t$, and measure the influence of $v$ on the prediction $\hgnn(v_t)$ w.r.t. $P$ by perturbing all the walks in $\mathcal{W}_G^P$. 
Henceforth, we may use $\explain{G}^P$ to denote a fine-grained explanation where $v$ and $v_t$ are the start and end nodes in $P$.

\subsection{Graph Rewiring for Walk Perturbation}

To perturb $\mathcal{W}_G^P$ for a fine-grained explanation $\explain{G}(v_t)=(v,P)$, we introduce a novel graph rewiring algorithm. 
The algorithm takes $G$ and $P$ as inputs and produces a rewired graph $G_R^P$ which guarantees the walks in $\mathcal{W}_G^P$ are blocked on $G_R^P$ and the other walks are reserved. 

Let $P$=$\langle v(v_0), v_1, \cdots, v_L, v_t(v_{L+1})\rangle$, $L\geq 0$.
The graph rewiring algorithm involves the following two steps.

$\bullet$ \emph{{Step 1: creating proxy nodes.}} It first creates a {proxy} node $\proxy(v_i)$ for every node $v_i$ ($i\in[1,L]$) in $P$. The feature vector and the type of each proxy node are the same as those of the original node. 
For ease of presentation, we let $v=\proxy(v)$ and $v_t=\proxy(v_t)$ though $v$ and $v_t$ do not have actual proxy nodes in the graph. 

$\bullet$ \emph{Step 2: establishing edges.}
The algorithm then establishes edges for the $L$ proxy nodes. 
For $v_i~(i\in[1,L])$, we denote by $\inedge(v_i)$ and $\outedge(v_i)$ the original sets of in-edges and out-edges of $v_i$ in $G$, respectively.
For $\langle u, v_i\rangle\in \inedge(v_i)$, if (i) it is a self-loop {\bf or} (ii) exactly one of the two edges $\langle u, v_i\rangle$, $\langle v_i, u\rangle$ exists in $P$, we add a new edge from $\proxy(u)$ to $\proxy(v_i)$. 
For $\langle v_i, u\rangle\in \outedge(v_i)$, if (i) it is not a self-loop {\bf and} (ii) both $\langle v_i, u\rangle$ and $\langle u,v_i\rangle$ are not contained in $P$, we add a new edge from $\proxy(v_i)$ to $u$.
Each newly added edge shares the same feature vector and the type of the original edge.
Finally, we remove the first edge $\langle v, v_1\rangle$ in $P$ from the graph.
\begin{algorithm}[t]		
	\caption{Rewiring algorithm}
	\label{alg:rewiring}
\textbf{Input:} $G$, $P=\langle v,v_1,\cdots, v_L,v_t\rangle$
\begin{algorithmic}[1]
\STATE $V\leftarrow V\cup \bigcup_{i=1}^L\{\proxy(v_i) \}$
\FOR{$i\in[1,L]$}
    \FOR{$\langle u, v_i \rangle \in \inedge(v_i)$}
        \IF{$ u=v_i \vee \langle u, v_i \rangle \in P \vee \langle v_i,u \rangle \in P$}
            \STATE $E\leftarrow E\cup \{\langle \proxy(u), \proxy(v_i)\rangle  \}$; 
        \ENDIF
    \ENDFOR
    \FOR{$\langle v_i, u \rangle\in \outedge(v_i)$}
        \IF{ $v_i\neq u \wedge \langle u, v_i \rangle \notin P \wedge \langle v_i,u \rangle \notin P$}
            \STATE $E\leftarrow E\cup \{\langle \proxy(v_i), u \rangle  \}$; 
        \ENDIF
    \ENDFOR
\ENDFOR   
\STATE $E\leftarrow E\backslash \{ \langle v, v_1 \rangle  \}$;
\STATE Return $G$;
	\end{algorithmic}

\end{algorithm}

The resultant graph is the rewired graph $G_R^P$. The overall process is formalized in Algorithm~\ref{alg:rewiring}. 

\subsubsection{Correctness Guarantee.}
We first use an example to show intuitions on why $G_R^P$ is able to block $\mathcal{W}_G^P$ without interfering with other walks. 
Without loss of generality, we assume every node and edge in $G$ has a distinct type. 
Informally, two walks are said to be \emph{equivalent} if their corresponding nodes and edges share the same feature vector and type. 
\begin{example}\label{ex:eq}
	Figure~\ref{fig:flowset-1} shows a path $P=\langle D, C, B, A\rangle$ in graph $G$. The rewired graph $G_R^P$ in Figure~\ref{fig:flowset-2} blocks $P$ by deleting $\langle D,C\rangle$ and disconnects $\proxy(B)$ from $A$. 
	Consider a walk $W_1=\langle D,C,B,C,B,A\rangle \in \mathcal{W}_G^P$. To mimic $W_1$ on $G_R^P$, we go from $D$ to $\proxy(C)$ but after that we can only walk between proxy nodes without having a chance to transit to $A$. This implies $W_1$ is blocked on $G_R^P$. 
	
	Consider two walks $W_2=\langle E, B, A\rangle$ and $W_3=\langle D, C, A \rangle$ which are not associated with $P$. It is clear that $W_2$ does not contain $\langle D, C\rangle$ and it still exists in $G_R^P$. For $W_3$, it contains an edge $e=\langle C, A\rangle$ after $\langle D, C\rangle$ which satisfies (i) $e$ is not a self-loop and (ii) both $e$ and its reverse edge are not contained in $P$. Hence, we can find a walk on $G_R^P$ that transits from $D$ to $\proxy(C)$ and goes back to $A$, i.e., $\langle D, \proxy(C), A \rangle$, which is equivalent to $W_3$.
	
\end{example}

%

\begin{definition}[Equivalent Walks]
	Let $W=\langle v_1, \cdots, v_L \rangle$ be a walk on $G$ and $W'=\langle v_1', \cdots, v_L' \rangle$ be a walk on $G_R^P$. 
	$W, W'$ are equivalent iff (i) for any $i\in[1,L]$, $v_i,v_i'$ have the same feature vector and the node type, and (ii) for any $i\in[1,L-1]$, $\langle v_i,v_{i+1}\rangle, \langle v_i',v_{i+1}'\rangle$ share the same feature vector and the edge type. We denote as $W=W'$. 
\end{definition}

Let $\mathcal{W}_G^{v_t}$ denote the set of all the walks on $G$ which end with node $v_t$. We have the following theorem.

\begin{theorem}
	For any fine-grained explanation $\explain{G}(v_t)=(v,P)$ for the prediction $\hgnn(v_t)$, Algorithm~\ref{alg:rewiring} produces a rewired graph $G_R^P$ that satisfies: 
	(i) $\mathcal{W}_G^P\cap \mathcal{W}_{G_R^P}^{v_t}=\emptyset$; and (ii) $\mathcal{W}_G^P\cup \mathcal{W}_{G_R^P}^{v_t}=\mathcal{W}_G^{v_t}$.
\end{theorem}

The above theorem conveys the correctness of our rewiring algorithm in two aspects. First, all the walks that are associated with $P$ in $G$ are blocked on $G_R^P$. Second, for any walk on $G$ which ends with $v_t$ and is not associated with $P$, we can find an equivalent walk on $G_R^P$.

\subsection{Fine-Grained Explainability for HGNs}

Before introducing our explanation framework, we first formulate the notion of \emph{influence score} of a fine-grained explanation. 
Given a fine-grained explanation $\explain{G}^P$ for the prediction $\hgnn(v_t)$, we use the rewired graph $G_R^P$ produced by Algorithm~\ref{alg:rewiring} to perturb all the walks in $\mathcal{W}_G^P$. We compute the model predictions on $v_t$ given $G$ and $G_R^P$, i.e., $M_G(v_t)$ and $M_{G_R^P}(v_t)$. The influence score of $\explain{G}^P$ is measured based on the significance of the prediction change.
\begin{definition}[Influence Score of $\explain{G}^P$]
	Let $M_G(v_t)$ and $M_{G_R^P}(v_t)$ denote the model predictions on $v_t$ given $G$ and $G_R^P$, respectively. The influence score of $\explain{G}^P$ is defined as:
	\begin{equation}\label{eq:sp}
	\small{
	{\emph\infl}(\explain{G}^P) = (-1)^{\mathbb{1}_{y=y'}} + (M_G(v_t)[y]-M_{G_R^P}(v_t)[y]),
}
	\end{equation}
	where $y=\arg\max_c M_G(v_t)$ and $y'=\arg\max_c M_{G_R^P}(v_t)$.
\end{definition}

The first term in Eq.~(\ref{eq:sp}) measures the change in the predicted label and the second term focuses on the change in the probability of the predicted label. A higher $\infl(\explain{G}^P)$ means blocking all the walks in $\mathcal{W}_G^P$ will change the prediction more significantly, indicating $\explain{G}^P$ is more critical to the prediction.
Moreover, if blocking the walks in $\mathcal{W}_G^P$ does not change the prediction label (i.e., $y$ equals $y'$), we have $\infl(\explain{G}^P)\in[-2,0]$. 
Otherwise, we have $\infl(\explain{G}^P)\in[0,2]$.
Note that $\infl(\explain{G}^P)<-1$ means blocking all the walks associated with $P$ actually increases the probability of the predicted label. We regard such explanations as invalid. 

\begin{algorithm}[t]		
	\caption{Finding top-$K$ fine-grained explanations}
    \label{alg:beam}
    \textbf{Input:} $v_t$, $\hgnn$, maximum iteration $L_{max}$, sample size $m$, candidate size $b$
    \begin{algorithmic}[1]
        \STATE $B\leftarrow\{\langle v_t \rangle\}$; $U\leftarrow \emptyset$; $i\leftarrow 0$; 
        \WHILE{$B\neq U$ and $i<L_{max}$}
        	\STATE $B\leftarrow U$; $U\leftarrow \emptyset$; $i\leftarrow i+1$;
            \FOR{$P\in B$ and $|P|=i$}
               \STATE Randomly sample $m$ one-step extended paths from $P$ and add them to $U$;

            \ENDFOR
            \STATE $U\leftarrow$ $b$ paths in $B\cup U$ with highest influence scores;
    	\ENDWHILE
        \STATE Return top-$K$ fine-grained explanations based on $B$;
    \end{algorithmic}  
\end{algorithm}
Given a prediction $\hgnn(v_t)$, our explanation framework aims to find $K$ fine-grained explanations with the highest influence scores. 
While each fine-grained explanation is exclusively specified by its simple path, i.e., the start node of the path is the cause for the prediction, it is intractable to enumerate all the simple paths ending with $v_t$ and compute their influence scores.
To avoid exhaustive search, we develop a greedy search algorithm, which is summarized in Algorithm~\ref{alg:beam}.
The core idea is to explore simple paths in order of their lengths.
We maintain a set $B$ of at most $b$ simple paths.
$B$ is initialized by a trivial path $\langle v_t\rangle$ of length zero, which only used to aid the algorithm but not considered as a real simple path.
At the $i$-th iteration ($i\in[0,L_{max})$), for each path of length $i$ in $B$, we randomly extend it by one-step $m$ times (based on the edges in $G$) and obtain $m$ paths of length $i+1$.
We use $U$ to maintain all the newly explored paths of length $i+1$.  
We then find at most $b$ paths in $B\cup U$ with the highest influence scores and update $B$ accordingly.
The iteration is repeated until all the paths in $B$ have been extended or the maximum iteration $L_{max}$ is reached, where $L_{max}$ can be set as the number of model layers.
Finally, we use the reverse of $K$ most-influential paths to form the top-$K$ fine-grained explanations. 

\subsubsection{Time Complexity Analysis.}
Algorithm~\ref{alg:beam} computes influence scores for at most $bmL_{max}$ fine-grained explanations. The time cost of graph rewiring is {\rm O}($pd$), where $p$ is the maximum length of paths in explanations and $d$ is the maximum node's degree in $G$.  The time cost of computing prediction change is {\rm O}($|\hgnn|$), where $|\hgnn|$ denotes the number of model parameters. 
Hence, the total time complexity of Algorithm~\ref{alg:rewiring} is {\rm O}($bmL_{max}(pd + |\hgnn|)$). 
In practice, the values of $b,m,L_{max},p$ are small constant numbers and $p\leq L_{max}$.  The empirical time cost of finding top-$K$ fine-grained explanations is linearly proportional to the model inference time.

\section{Experiments}
\eat{
In this section, we conduct experiments to answer the following research questions:
\begin{itemize}
    \item \textbf{RQ1} How is the effectiveness of our proposed \modelname for explaining HGNs compared with the state-of-the-art explanation methods?

    \item \textbf{RQ2} How is the efficiency of \modelname in finding fine-grained explanations for HGNs?

    \item \textbf{RQ3} Are the fine-grained explanations found by \modelname user-friendly?
\end{itemize}
}

\subsection{Experimental Settings}
\subsubsection{Datasets.}
We conduct experiments using three public heterogeneous graph datasets for node classification tasks. 
(1) \textbf{ACM}~\cite{wang2021self} is an academic network with node types: \textit{paper} (P), \textit{author} (A) and \textit{subject} (S). The task is to classify papers into three topics.
(2) \textbf{DBLP}~\cite{wang2021self} is a bibliography graph with node types: \textit{author} (A), \textit{paper} (P), \textit{conference} (C) and \textit{term} (T). The task is to classify authors into four research areas.
(3) \textbf{IMDB}\footnote{\url{https://www.kaggle.com/carolzhangdc/imdb-5000-movie-dataset}} is a movie graph with node types: \textit{movie} (M), \textit{director} (D), and \textit{actor} (A). 
The statistics of the datasets are in Table~\ref{tab:dataset}.
\subsubsection{HGN Models and Training Details.}
We evaluate the performance of explanation methods on three advanced HGNs: a 2-layer SimpleHGN~\cite{lv2021we}, a 3-layer SimpleHGN and a 2-layer HGT~\cite{hu2020heterogeneous}, which are abbreviated as {\bf SIM2}, {\bf SIM3} and {\bf HGT}. 
The node embedding size is set to 32 in all the models. 
To train these models, we split each dataset into training/validation/test sets and we randomly reserve 1000/1000 samples for validation/test. Note that HGT contains more parameters and needs more training samples to reach good performance. While SimpleHGN is parameter-efficient, its performance increases insignificantly with more training samples. We train HGT with 2000 samples and train SimpleHGN with 60 samples per label for training efficiency. Table~\ref{tab:dataset} provides the test accuracy.

\subsubsection{Compared Explanation Methods.}
Existing explainability techniques are proposed for GNNs and there is no off-the-self ground-truth explanations for the predictions from HGNs. Hence, we implement two basic explanation methods and adapt four state-of-the-art model-agnostic GNN explanation approaches as comparison methods. 
(1) \textbf{Local} only considers graph structure and selects nodes with the shortest distances to the target node. It is task-agnostic and generates the same explanation for all the HGNs.
(2) \textbf{Attention} utilizes the attention scores computed by HGNs during message aggregation. 
For a node, we consider all the paths to the target node. We multiply the attention scores of the edges in each path, and add them up as node importance to the prediction. It is a model-specific explanation method. 
(3) \textbf{PGM-Explainer}~\cite{vu2020pgm} perturbs node features to construct a dataset characterizing local data distribution and generates node-level explanations in form of PGMs. To adapt it to HGNs, we perturb a node by replacing its feature vector with the average feature vector of the nodes with the same type.
%
(4) \textbf{ReFine}~\cite{wang2021towards} is a learning-based method which provides edge-level explanations. 
It uses a graph encoder followed by an MLP layer to produce the probability of an edge in the explanation, and trains the model using fidelity loss and contrastive loss. We use HGNs to instantiate the graph encoder. 
%
(5) \textbf{Gem}~\cite{lin2021generative} provides edge-level explanations. It involves a distillation process to obtain ground-truth explanations in advance and trains a model that predicts the probabilities of edges in the explanations. We use HGNs to implement the node embedding module. 
%
(6) \textbf{SubgraphX}~\cite{yuan2021explainability} provides subgraph-level explanations. It performs the Monte Carlo tree search to explore different subgraphs and measures their importance. To adapt it to HGNs, we simply ignore all the type information during the search process. 
\begin{table}[t]
    \setlength\tabcolsep{5pt}
	\centering
	\small
	\begin{tabular}{c|cccc|ccc}
		\toprule
		\!\!\!\!\!\!{Dataset}\!\!\!\!\!\! &\!\!\! \#nodes \!\!\!& \!\!\!\makecell[c]{\#node\\types} \!\!\!& \!\!\!\#edges \!\!\! &\!\!\!\makecell[c]{\#edge\\types} \!\!\!& \!\!\!SIM2\!\!\! & \!\!\!SIM3\!\!\! & \!\!\!HGT\!\!\!\\
		\hline
		ACM &11246 &3 &46098 &7 & 90.4 & 91.3 & 89.7  \\
		DBLP &26128 &4 &265694 &10 & 95.8 & 95.5 & 87.3\\
		IMDB &11616 &3 &45828 &7 &61.6  &61.1 & 72.4\\
		\bottomrule
	\end{tabular}
        \caption{The statistics of the datasets and the test accuracy(\%) of the HGNs.}
	\label{tab:dataset}
\end{table}

\begin{table*}[t!]
	\centering
	
	\small
	\begin{tabular}{c|c|c|ccccccc}
		\toprule
		Dataset    & Metric    & Model     & Local     & Attention          & PGM-Explainer     & ReFine                & GEM               & SubgraphX                  & \modelname  \\ \hline
		\multirow{6}*{ACM} & \multirow{3}*{$F^{acc}$} 
		& SIM2   & {\ul 2.7} & 4.2           & 18.1\std{0.31}                      & {17.2\std{1.25}}        & 4.2\std{0.00}       & {3.8\std{0.49}}             & \textbf{0.2$^\ast$\std{0.09}}\\ 
		~           & ~         & SIM3    & 5.4       & 6.5           & 31.7\std{0.34}                      & {26.0\std{0.14}}          & 26.9\std{0.01}      & {{\ul 4.6\std{0.20}}}       & \textbf{4.1\std{0.23}}   \\            
		~           & ~         & HGT     &  {\ul{0.7}}       & {\textbf{0.2}}     & 46.7\std{0.52}                      & {4.2\std{1.67}}         & 1.9\std{0.00}       &  1.4\std{0.33}    & 5.5\std{0.19}  \\ \cline{2-10}            
		~ & \multirow{3}*{$F^{prob}$} 
		& SIM2    & 0.7       & {1.2}     & 6.8\std{0.08}                       & {6.2\std{0.31}}         & {1.2\std{0.00}} & {\ul 0.6\std{1.23}}             & \textbf{-0.9$^\ast$\std{0.02}}  \\             
		~           & ~         & SIM3    & 5.4       & 2.4           & 11.4\std{0.02}                      & {{10.9\std{0.05}}}  & 11.3\std{0.00}      & {\ul 1.5\std{0.31}}             & \textbf{0.7$^\ast$ \std{0.09}} \\             
		~           & ~         & HGT     & {\ul 0.6} & \textbf{0.003}  & 35.2\std{0.34}                      & {3.3\std{0.99}}         & 1.6\std{0.00}       & {1.0\std{0.07}}             & 1.9\std{0.30} \\ \hline            
		\multirow{6}*{DBLP} &\multirow{3}*{$F^{acc}$} 
		& SIM2    & 12.1      & 6.8           & 2.3\std{0.15}                       & {4.7\std{0.10}}         & 11.4\std{0.15}      & {{\ul 0.7\std{0.57}}}       & \textbf{0.6\std{0.09}} \\            
		~           & ~         & SIM3    & 7.0       & {\ul 4.9}     & 31.9\std{0.26}                      & {14.5\std{0.75}}        & 32.1\std{0.00}       & {5.9\std{0.05}}             & \textbf{0.3$^\ast$\std{0.05}}  \\            
		~           & ~         & HGT     & 3.0       & 2.5           & 6.2\std{0.24}                       & {7.7\std{1.12}}         & 6.1\std{0.00}       & {{\ul 0.8\std{0.06}}}       & \textbf{0.3$^\ast$\std{0.00}} \\ \cline{2-10}             
		~ & \multirow{3}*{$F^{prob}$} 
		& SIM2    & 4.2       & 2.2           & 0.8\std{0.05}                       & {0.9\std{0.10}}         & 3.8\std{0.00}       & {\textbf{-0.8\std{3.43}}}   & {\ul -0.5\std{0.01}} \\              
		~           & ~         & SIM3    & 2.7       & {\ul 1.7}     & 11.3\std{0.03}                      & {5.3\std{0.67}}         & 11.4\std{0.01}      & {2.1\std{0.31}}             & \textbf{-0.4$^\ast$ \std{0.01}} \\               
		~           & ~         & HGT     & 3.9       & 3.5           & 5.9\std{0.22}                       & {6.2\std{0.68}}         & 6.3\std{0.00}       & {\textbf{-2.3\std{0.07}}}   & {\ul -1.0\std{0.03}} \\ \hline            
		\multirow{6}*{IMDB} & \multirow{3}*{$F^{acc}$} 
		& SIM2    & 6.3       & 6.3           & 6.0\std{0.93}                         & {13.4\std{0.81}}        & 18.8\std{8.49}      & {{\ul 2.4\std{0.38}}}       & \textbf{0.9$^\ast$\std{0.08}} \\          
		~           & ~         & SIM3    & 10.1      & 10.1          & 9.4\std{1.15}                       & {16.8\std{0.82}}        & 17.1\std{1.20}      & {{\ul 2.8\std{0.05}}}       & \textbf{2.5\std{0.00}}  \\           
		~           & ~         & HGT     & {\ul 0.1} & {\ul 0.1}     &3.4\std{0.68}                       & {4.9\std{0.66}}         & 13.5\std{2.20}      & {1.4\std{0.02}}             & \textbf{0.0\std{0.00}} \\ \cline{2-10}            
		~ & \multirow{3}*{$F^{prob}$} 
		& SIM2    & 1.3       & 1.3           & 1.0\std{0.20}                       & {3.1\std{0.15}}         & 5.1\std{2.68}       & {{\ul -1.7\std{2.31}}}      & \textbf{-2.8\std{0.01}}\\          
		~           & ~         & SIM3    & 2.3       & 2.3           & 2.0\std{0.57}                       & {4.8\std{0.45}}         & 5.1\std{0.20}       & {{\ul -1.6\std{0.51}}}      & \textbf{-2.0$^\ast$\std{0.01}} \\           
		~           & ~         & HGT     & \small{\ul 0.02} & \small{\ul 0.02}     & 4.8\std{0.22}                       & {4.0\std{0.24}}         & 13.3\std{2.57}      & {0.1\std{0.14}}             & \textbf{-2.8$^\ast$\std{0.00}} \\
		\bottomrule
	\end{tabular}
	
        \caption{Comparison results on fidelity metrics(\%). The best results 
		are in bold and the second-best results are underlined. 
            And  $^\ast$ denotes statistically significant improvement (measured by t-test with p-value $< 0.01$) over all baselines.
	}
     \label{tab:faith}
\end{table*}

\subsubsection{Evaluation Metrics.}

A faithful explanation should involve graph objects (e.g., nodes, edges, subgraphs) that are necessary and sufficient to recover the original prediction.
Without ground-truth explanations, we follow the previous work~\cite{yuan2020explainability} and measure faithfulness by two fidelity metrics: 
the \emph{accuracy fidelity} $F^{acc}$ measures the prediction change and the \emph{probability fidelity} $F^{prob}$ studies the probability change on the original predicted label.  
Specifically, let $\mathcal{T}$ denote test samples correctly predicted by the model. For a test sample ($v_t, y$), we induce a graph $\tilde{G}$ based on all the involving graph objects in its explanation. We then compute the prediction $M_{\tilde G}(v_t)$ and the predicted label $\tilde{y}$ based on $\tilde{G}$.
Formally, we compute metrics by:
{\small
\begin{align}
F^{acc}&= \frac{1}{|\mathcal{T}|}\sum_{(v_t,y)\in\mathcal{T}} (1-\mathbb{1}_{y=\Tilde{y}}),\\
F^{prob}&= \frac{1}{|\mathcal{T}|}\sum_{(v_t,y)\in\mathcal{T}} (M_G(v_t)[y]-M_{\tilde{G}}(v_t)[y]).
\end{align}
}

\noindent\emph{Remark.} \modelname focuses on evaluating the influence of a node on the prediction w.r.t. a specific path. 
It is unreasonable to regard the reported top-$K$ fine-grained explanations with the highest influence scores as a subgraph composed of $K$ paths.
This is because our rewiring algorithm only perturbs the walks that are associated with the path in a fine-grained explanation, rather than treating the path as a subgraph and excluding all the involving graph objects from the graph.
However, the two fidelity metrics actually treat our top-$K$ fine-grained explanations as a subgraph explanation and measure the effectiveness of the subgraph induced by the paths. In this regard, the two metrics do not act fairly to \modelns.

\subsubsection{Implementation Details.}
We implemented \modelname with PyTorch. For the four advanced explanation approaches, we used their original source codes and incorporated simple adaptations as described before to make them fit heterogeneous graphs.
For \modelns, we tune the hyperparameters $(b,m)$ of ACM to $(5,5)$, DBLP with SIM2 to $(10,10)$, and others to $(2,10)$.
Since explanation sparsity is highly related to fidelity scores, we control the number of nodes involved in the explanations. 
As suggested by the previous work~\cite{ying2019gnnexplainer}, we set the node number to 5 to avoid overwhelming users. 
As the comparison methods may not find explanations involving exactly 5 nodes, we tune their hyperparameters to find explanations with roughly 5 nodes that achieve the best fidelity.
We conducted all the experiments on a server equipped with Intel(R) Xeon(R) Silver 4110 CPU, 128GB Memory, and a Nvidia GeForce RTX 2080 Ti GPU (12GB Memory). 
Each experiment was repeated 5 times and the average performance was reported.

\subsection{Evaluation of Effectiveness}

\subsubsection{Comparison Results.}
\label{section:faith}

Table~\ref{tab:faith} shows the fidelity performance of all the comparison methods.
On IMDB, Local and Attention have the same results because these basic methods give the same explanations. 
For both metrics, lower fidelity scores indicate the graph objects in the explanations are more useful to retain the original predictions, and hence the explanations are more faithful to the model. We have the following key observations. 
First, \modelname achieves the lowest fidelity scores in most cases, demonstrating its effectiveness in identifying explanations that are faithful to the models. 
Interestingly, \modelname reports negative probability fidelity values in some cases, which indicates \modelname is capable of filtering out noisy information and identifying information flows that are critical to the predictions.  
When explaining HGT on ACM dataset, \modelname performs worse than Attention. 
We investigate the test samples in ACM where the explanations from \modelname fail to provide the original predictions, and find that the paths in these explanations do not contain nodes of the \emph{author} type. 
This is because the \emph{subject} nodes have much larger degrees than the \emph{author} nodes and the paths following \emph{paper-subject-paper} dominate all the paths ending with a target paper. 
If many of these paths are indeed influential to HGT, they will occupy the top-K position, making top-$K$ explanations lack \textit{author} type information.
To verify our claim, for each test sample in ACM, 
we look up paths examined during the search process for those following \emph{author-paper} and \emph{paper-author-paper}, and use the one with the highest influence score to replace the $K$-th path identified by \modelns. 
This simple strategy achieves better results ($F^{acc}$=0.0\% and $F^{prob}$=-2.0\%) than Attention.
This indicates \modelname is extensible to explore the correlations among explanations towards better fidelity. 
On probability fidelity, \modelname performs slightly worse than SubgraphX on DBLP. This is because the influence score is defined to be less sensitive to the change in probabilities and \modelname pays more attention to avoiding label change, i.e., better accuracy fidelity than SubgraphX.
Second, among the adaptations of four advanced explanation methods, SubgraphX is superior in all cases. This is consistent with the intuition that high-order explanations(e.g., subgraphs) are more expressive and powerful than low-order ones(e.g., nodes, edges).
PGM-Explainer is sensitive to graph data. Its fidelity scores increase sharply on ACM and DBLP than IMDB because the node neighborhoods in ACM and DBLP is larger, which increases the difficulty of obtaining accurate local data distributions via sampling-based perturbations. 
GEM and ReFine fail to find good explanations, which suggests that edge-based explanations are insufficient to retain the rich semantics that are useful to the HGN predictions. 
We also notice that Attention is a strong baseline thanks to the knowledge of model details. Local provides task-agnostic explanations and its performance is unstable over different model architectures.
Third, \modelname performs steadily on different HGNs and tasks. On DBLP, the four existing explanation methods report much higher accuracy fidelity on SIM3 than on SIM2. 
We notice that SIM3 involves larger computation graphs than SIM2 (about two orders of magnitude in the number of nodes), and existing explanation methods fall short in the presence of intricate graph structures. 
\modelname achieves best fidelity when explaining SIM3 on all three tasks, showing the potential of applying \modelname on complex HGNs and large graphs.  

\subsubsection{Effectiveness of Influence Scores.} 
\begin{figure}[t]
        \centering
        \begin{subfigure}{0.47\linewidth}
            \includegraphics[width=\linewidth]{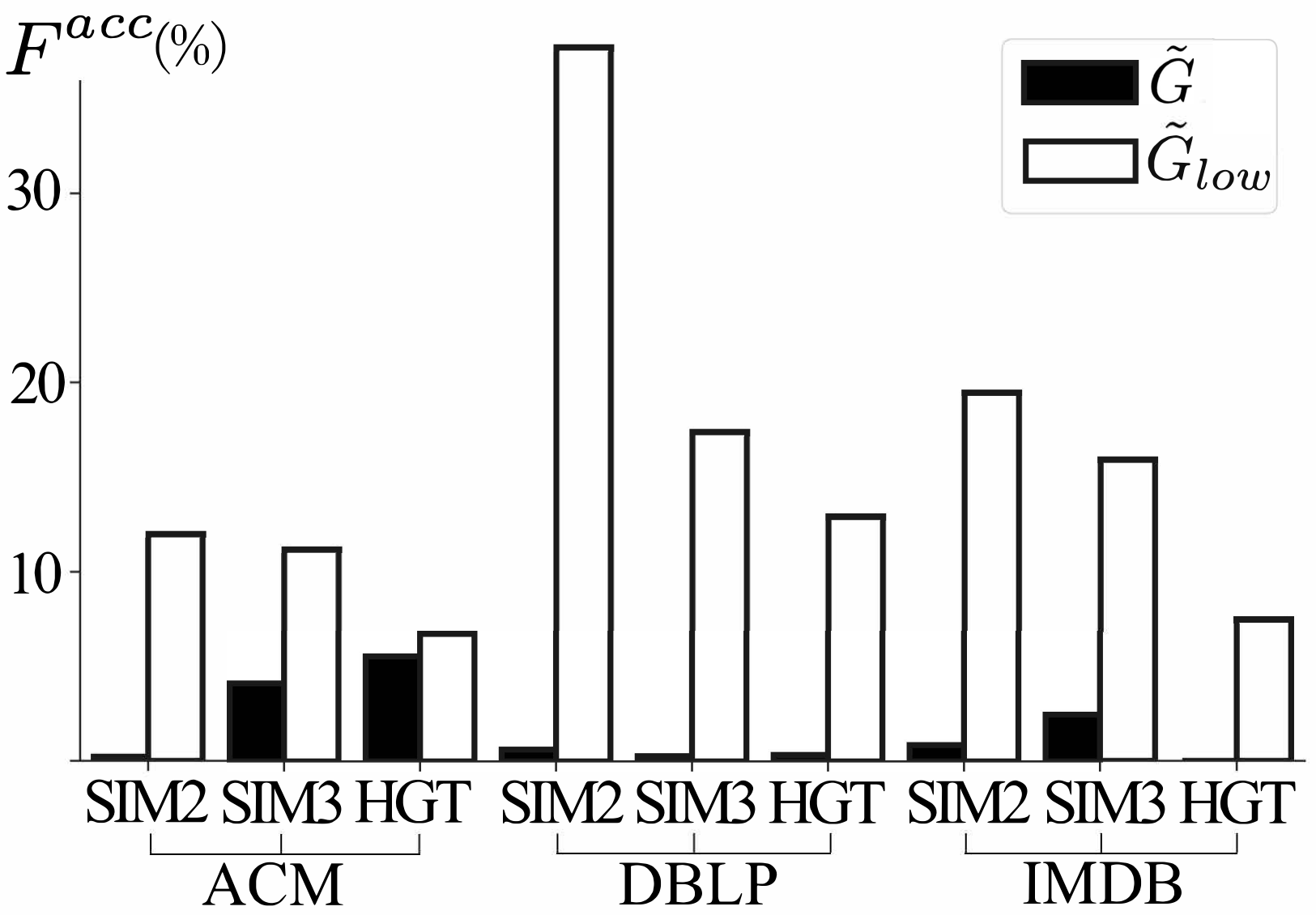}
            \caption{Accuracy fidelity}
            \label{fig:level-facc}
        \end{subfigure}
		\quad
        \begin{subfigure}{0.47\linewidth}
            \includegraphics[width=\linewidth]{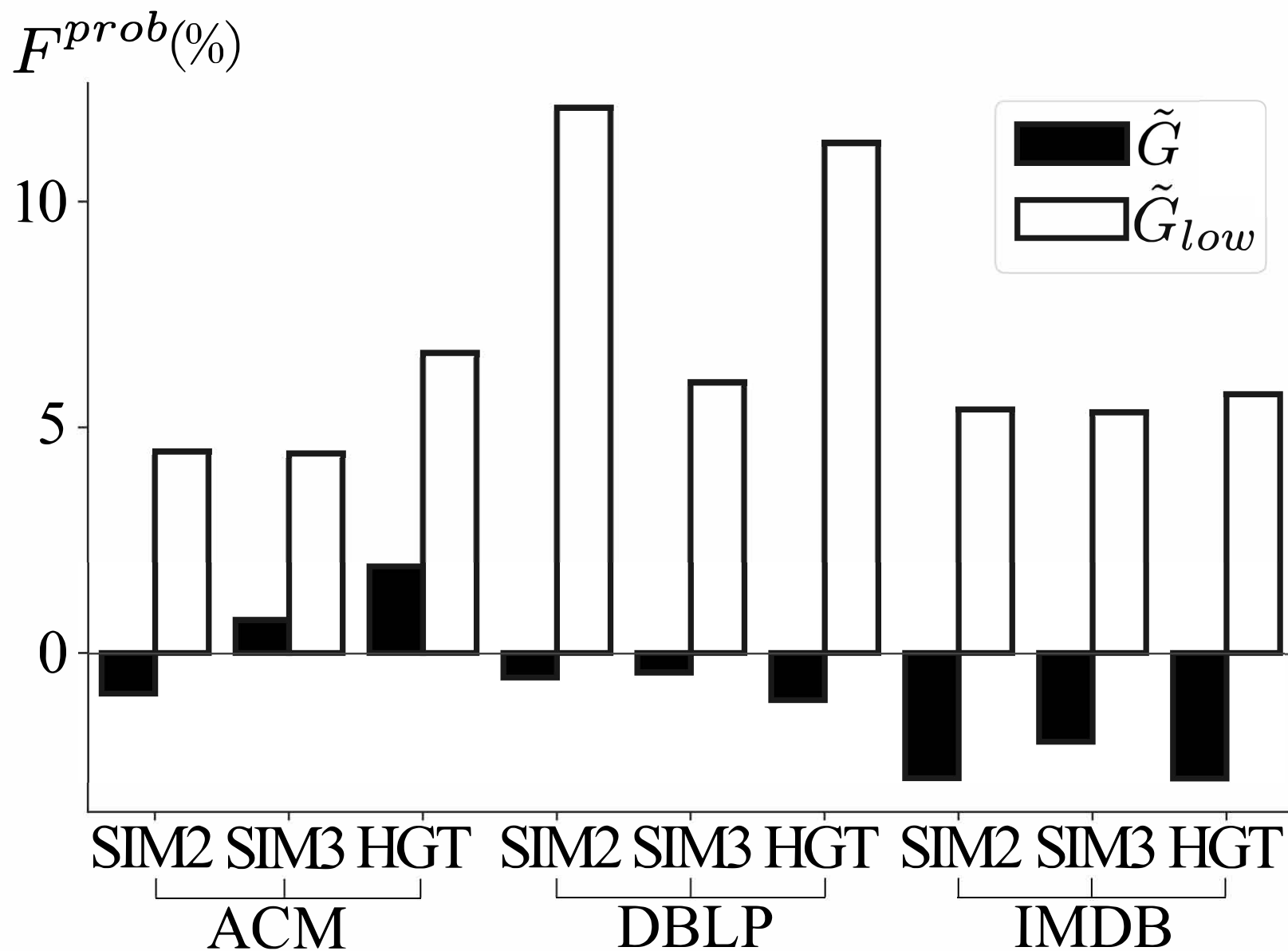}
            \caption{Probability fidelity}
            \label{fig:level-fprob}
        \end{subfigure}
        \caption{Effectiveness of influence scores in differentiating faithful explanations from unfaithful ones.}
        \label{fig:level-verify}
\end{figure}

We study whether proposed influence score (Eq.~\ref{eq:sp}) can differentiate faithful explanations from unfaithful ones. 
We consider all the paths examined during the search process and induce another graph $\tilde{G}_{low}$ based on the $K$ paths with the lowest influence scores.
Figure~\ref{fig:level-verify} provides the fidelity scores computed based on $\tilde{G}$ and $\tilde{G}_{low}$. 
In all the cases, $\tilde{G}_{low}$ consistently result in poor fidelity. The performance gap between two groups confirms the effectiveness of the influence score function in distinguishing different fine-grained explanations. 



\subsection{Evaluation of Efficiency}
We report the running time in computing explanations for all the test samples in Table~\ref{tab:eff}. 
\modelname is much more efficient than the existing explanation methods, thanks to the greedy search algorithm. 
The sampling process in PGM-Explainer and the Monte Carlo search in SubgraphX incur high time costs when the target node has a large neighborhood, e.g., SIM3 on DBLP. GEM is efficient in many cases but also suffers in explaining SIM3 due to large node neighborhood, because it calculates the probability of edges between pairwise nodes in the neighborhood. 
ReFine generally runs faster than PGM-Explainer and SubgraphX. 
Note that the time cost of the learning-based methods GEM and ReFine mainly comes from training the explanation generators and the time for finding explanations for test samples can be amortized.

\begin{table}[t]
	\centering
	
	\small
	\begin{tabular}{c|c|ccccc}
		\toprule
		\!\!\!{Dataset}\!\!\!& \!\!\!{Model}\!\!\!& \!{\makecell[c]{PGM-\\Explainer}}\!\!&\!{ReFine}\!\!\!            &\!\!\!{GEM}\!\!\! & \!\makecell[c]{Sub-\\graphX}\!\!\! &\! {\modelname} \!\!\!\!\!  \\ \hline
		\multirow{3}{*}{ACM}  & SIM2 &  19.6       &  4.6  &  2.6  &  3.5    &  0.2 \\
		& SIM3 &  18.6        &  5.8  &  45.5 &  13.5    &  0.6 \\
		& HGT  &  16.7        &  3.6  &  2.6  &  3.9    &  0.2 \\ \hline
		\multirow{3}{*}{DBLP} & SIM2 &  12.4        &  5.9  &  9.7  &  7.5    &  0.6 \\
		& SIM3 &  46.0       &  11.0 &  84.5 &  27.9   &  1.2 \\
		& HGT  &  11.3        &  4.7 &  3.3 &  8.0    &  0.2 \\ \hline
		\multirow{3}{*}{IMDB} & SIM2 &  3.1        &  3.0  &  1.5 &  2.6    &  0.1 \\
		& SIM3 &  10.2        &  4.8  &  22.6 &  9.0    &  0.1 \\
		& HGT  &  5.5        &  3.5  &  3.0  &  5.2    &  0.1 \\ \bottomrule
	\end{tabular}
	
        \caption{The time of computing explanations (in hours).}
        \label{tab:eff}
\end{table}

\subsection{Case Study}
\begin{figure}[t]
        \centering
        \begin{subfigure}[b]{0.48\columnwidth}
            \includegraphics[width=\textwidth]{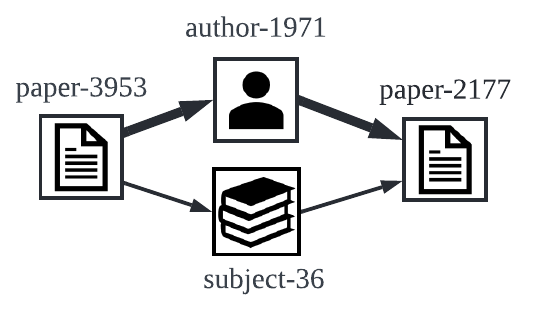}
            \caption{}
            \label{fig:case-1}
        \end{subfigure}
        \hfill
        \begin{subfigure}[b]{0.48\columnwidth}
            \includegraphics[width=\textwidth]{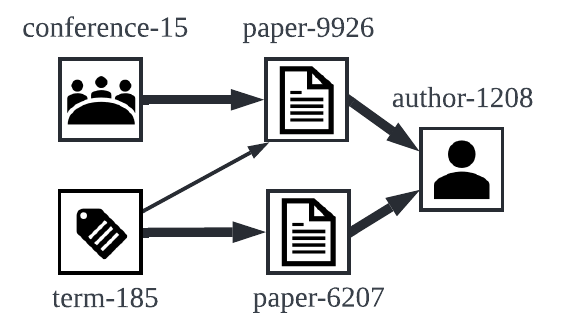}
            \caption{}
            \label{fig:case-2}
        \end{subfigure}
        \caption{Case studies: (a) explaining SIM3 on \textit{paper}-2177 in ACM; (b) explaining SIM3 on \textit{author}-1208 in DBLP.}
        \label{fig:case}
\end{figure}

We conduct two case studies to provide intuitions on how \modelname answers the ``what'' and ``how'' questions via fine-grained explanations. 
%
As shown in Figure~\ref{fig:case-1}, when explaining the prediction of \textit{paper}-2177, \modelname identifies two influential fine-grained explanations with paths $\langle$\textit{paper}-3953, \textit{author}-1971, \textit{paper}-2177$\rangle$ and $\langle$\textit{paper}-3953, \textit{subject}-36, \textit{paper}-2177$\rangle$, whose influence scores are 1.04 and -1.00, respectively. This means the model made the prediction mainly because \textit{paper}-2177, \textit{paper}-3953 are written by the same author. 
Figure~\ref{fig:case-2} shows the neighborhood of the target node \textit{author}-1208 in DBLP.
\modelname finds two nodes \textit{conference}-15 and \textit{term}-185 that are important for the prediction. The most influential path for \textit{conference}-15 is $\langle$\textit{conference}-15, \textit{paper}-9926, \textit{author}-1208$\rangle$, while that for \textit{term}-185 is $\langle$\textit{term}-185, \textit{paper}-6207, \textit{author}-1208$\rangle$. 
\modelname highlights two cause nodes and their influential paths, indicating the model made the prediction of \textit{author}-1208 because the author has written \textit{paper}-9926 in \textit{conference}-15 and published \textit{paper}-6207 involving \textit{term}-185.
Arguably, \modelname can provide explanations with legible semantics, which is a desirable property in explaining HGNs on complex heterogeneous graphs.

\section{Conclusion}

In this paper, we study the problem of explaining black-box HGNs on node classification tasks. 
We propose a new explanation framework named \modelname which provides fine-grained explanations in the form of a node associated with its influence path to the target node. The node tells what is important to the prediction and the influence path indicates how the prediction is affected by the node. 
In \modelns, we develop a novel graph rewiring algorithm to perform walk-level perturbation and measure the influence score of any fine-grained explanation without the knowledge of model details. 
We further introduce a greedy search algorithm to find top-$K$ most influential explanations efficiently. Extensive experimental results show that \modelname can provide explanations that are faithful to various HGNs with high efficiency, outperforming the adaptations of known explainability techniques.

\section{Acknowledgements}
The authors would like to thank the anonymous reviewers for their insightful reviews and the deep learning computing framework MindSpore\footnote{https://www.mindspore.cn/} for the support on this work. This work is supported by the National Key Research and Development Program of China (2022YFE0200500), Shanghai Municipal Science and Technology Major Project (2021SHZDZX0102) and SJTU Global Strategic Partnership Fund (2021 SJTU-HKUST).

\bibliography{6437.LiT.bib}

\begin{thebibliography}{21}
\providecommand{\natexlab}[1]{#1}

\bibitem[{Baldassarre and Azizpour(2019)}]{baldassarre2019explainability}
Baldassarre, F.; and Azizpour, H. 2019.
\newblock Explainability Techniques for Graph Convolutional Networks.
\newblock In \emph{International Conference on Machine Learning (ICML)
  Workshops, 2019 Workshop on Learning and Reasoning with Graph-Structured
  Representations}.

\bibitem[{Dou et~al.(2020)Dou, Liu, Sun, Deng, Peng, and Yu}]{dou2020enhancing}
Dou, Y.; Liu, Z.; Sun, L.; Deng, Y.; Peng, H.; and Yu, P.~S. 2020.
\newblock Enhancing graph neural network-based fraud detectors against
  camouflaged fraudsters.
\newblock In \emph{Proceedings of the 29th ACM International Conference on
  Information \& Knowledge Management}, 315--324.

\bibitem[{Fu et~al.(2020)Fu, Zhang, Meng, and King}]{fu2020magnn}
Fu, X.; Zhang, J.; Meng, Z.; and King, I. 2020.
\newblock Magnn: Metapath aggregated graph neural network for heterogeneous
  graph embedding.
\newblock In \emph{Proceedings of The Web Conference 2020}, 2331--2341.

\bibitem[{Hu et~al.(2020)Hu, Dong, Wang, and Sun}]{hu2020heterogeneous}
Hu, Z.; Dong, Y.; Wang, K.; and Sun, Y. 2020.
\newblock Heterogeneous graph transformer.
\newblock In \emph{Proceedings of The Web Conference 2020}, 2704--2710.

\bibitem[{Li et~al.(2021)Li, Peng, Cao, Dou, Zhang, Yu, and He}]{li2021higher}
Li, J.; Peng, H.; Cao, Y.; Dou, Y.; Zhang, H.; Yu, P.; and He, L. 2021.
\newblock Higher-order attribute-enhancing heterogeneous graph neural networks.
\newblock \emph{IEEE Transactions on Knowledge and Data Engineering}.

\bibitem[{Lin, Lan, and Li(2021)}]{lin2021generative}
Lin, W.; Lan, H.; and Li, B. 2021.
\newblock Generative causal explanations for graph neural networks.
\newblock In \emph{International Conference on Machine Learning}, 6666--6679.
  PMLR.

\bibitem[{Liu et~al.(2021)Liu, Ao, Qin, Chi, Feng, Yang, and He}]{liu2021pick}
Liu, Y.; Ao, X.; Qin, Z.; Chi, J.; Feng, J.; Yang, H.; and He, Q. 2021.
\newblock Pick and choose: a GNN-based imbalanced learning approach for fraud
  detection.
\newblock In \emph{Proceedings of the Web Conference 2021}, 3168--3177.

\bibitem[{Luo et~al.(2020)Luo, Cheng, Xu, Yu, Zong, Chen, and
  Zhang}]{luo2020parameterized}
Luo, D.; Cheng, W.; Xu, D.; Yu, W.; Zong, B.; Chen, H.; and Zhang, X. 2020.
\newblock Parameterized explainer for graph neural network.
\newblock \emph{Advances in neural information processing systems}, 33:
  19620--19631.

\bibitem[{Lv et~al.(2021)Lv, Ding, Liu, Chen, Feng, He, Zhou, Jiang, Dong, and
  Tang}]{lv2021we}
Lv, Q.; Ding, M.; Liu, Q.; Chen, Y.; Feng, W.; He, S.; Zhou, C.; Jiang, J.;
  Dong, Y.; and Tang, J. 2021.
\newblock Are we really making much progress? Revisiting, benchmarking and
  refining heterogeneous graph neural networks.
\newblock In \emph{Proceedings of the 27th ACM SIGKDD Conference on Knowledge
  Discovery \& Data Mining}, 1150--1160.

\bibitem[{Pope et~al.(2019)Pope, Kolouri, Rostami, Martin, and
  Hoffmann}]{pope2019explainability}
Pope, P.~E.; Kolouri, S.; Rostami, M.; Martin, C.~E.; and Hoffmann, H. 2019.
\newblock Explainability methods for graph convolutional neural networks.
\newblock In \emph{Proceedings of the IEEE/CVF Conference on Computer Vision
  and Pattern Recognition}, 10772--10781.

\bibitem[{Schlichtkrull et~al.(2018)Schlichtkrull, Kipf, Bloem, Berg, Titov,
  and Welling}]{schlichtkrull2018modeling}
Schlichtkrull, M.; Kipf, T.~N.; Bloem, P.; Berg, R. v.~d.; Titov, I.; and
  Welling, M. 2018.
\newblock Modeling relational data with graph convolutional networks.
\newblock In \emph{European semantic web conference}, 593--607. Springer.

\bibitem[{Schlichtkrull, De~Cao, and
  Titov(2020)}]{schlichtkrull2020interpreting}
Schlichtkrull, M.~S.; De~Cao, N.; and Titov, I. 2020.
\newblock Interpreting Graph Neural Networks for NLP With Differentiable Edge
  Masking.
\newblock In \emph{International Conference on Learning Representations}.

\bibitem[{Schnake et~al.(2021)Schnake, Eberle, Lederer, Nakajima, Schutt,
  Mueller, and Montavon}]{schnake2021higher}
Schnake, T.; Eberle, O.; Lederer, J.; Nakajima, S.; Schutt, K.~T.; Mueller,
  K.-R.; and Montavon, G. 2021.
\newblock Higher-Order Explanations of Graph Neural Networks via Relevant
  Walks.
\newblock \emph{IEEE Transactions on Pattern Analysis \& Machine Intelligence},
  (01): 1--1.

\bibitem[{Vu and Thai(2020)}]{vu2020pgm}
Vu, M.; and Thai, M.~T. 2020.
\newblock Pgm-explainer: Probabilistic graphical model explanations for graph
  neural networks.
\newblock \emph{Advances in neural information processing systems}, 33:
  12225--12235.

\bibitem[{Wang et~al.(2022)Wang, Bo, Shi, Fan, Ye, and Philip}]{wang2022survey}
Wang, X.; Bo, D.; Shi, C.; Fan, S.; Ye, Y.; and Philip, S.~Y. 2022.
\newblock A survey on heterogeneous graph embedding: methods, techniques,
  applications and sources.
\newblock \emph{IEEE Transactions on Big Data}.

\bibitem[{Wang et~al.(2021{\natexlab{a}})Wang, Liu, Han, and
  Shi}]{wang2021self}
Wang, X.; Liu, N.; Han, H.; and Shi, C. 2021{\natexlab{a}}.
\newblock Self-supervised heterogeneous graph neural network with
  co-contrastive learning.
\newblock In \emph{Proceedings of the 27th ACM SIGKDD Conference on Knowledge
  Discovery \& Data Mining}, 1726--1736.

\bibitem[{Wang et~al.(2021{\natexlab{b}})Wang, Wu, Zhang, He, and
  Chua}]{wang2021towards}
Wang, X.; Wu, Y.; Zhang, A.; He, X.; and Chua, T.-S. 2021{\natexlab{b}}.
\newblock Towards multi-grained explainability for graph neural networks.
\newblock \emph{Advances in Neural Information Processing Systems}, 34:
  18446--18458.

\bibitem[{Yang et~al.(2021)Yang, Guan, Li, Zhao, Cui, and
  Wang}]{yang2021interpretable}
Yang, Y.; Guan, Z.; Li, J.; Zhao, W.; Cui, J.; and Wang, Q. 2021.
\newblock Interpretable and efficient heterogeneous graph convolutional
  network.
\newblock \emph{IEEE Transactions on Knowledge and Data Engineering}.

\bibitem[{Ying et~al.(2019)Ying, Bourgeois, You, Zitnik, and
  Leskovec}]{ying2019gnnexplainer}
Ying, Z.; Bourgeois, D.; You, J.; Zitnik, M.; and Leskovec, J. 2019.
\newblock Gnnexplainer: Generating explanations for graph neural networks.
\newblock \emph{Advances in neural information processing systems}, 32.

\bibitem[{Yuan et~al.(2020)Yuan, Yu, Gui, and Ji}]{yuan2020explainability}
Yuan, H.; Yu, H.; Gui, S.; and Ji, S. 2020.
\newblock Explainability in graph neural networks: A taxonomic survey.
\newblock \emph{arXiv preprint arXiv:2012.15445}.

\bibitem[{Yuan et~al.(2021)Yuan, Yu, Wang, Li, and Ji}]{yuan2021explainability}
Yuan, H.; Yu, H.; Wang, J.; Li, K.; and Ji, S. 2021.
\newblock On explainability of graph neural networks via subgraph explorations.
\newblock In \emph{International Conference on Machine Learning}, 12241--12252.
  PMLR.

\end{thebibliography}

\end{document}